\pdfoutput=1

\documentclass[11pt]{article}
\usepackage{multirow}
\usepackage{booktabs}
\usepackage{graphicx}
\usepackage{cuted}  
\usepackage{xcolor}
\usepackage{tcolorbox} 
\usepackage{float}
\usepackage[preprint]{acl}

\usepackage{times}
\usepackage{latexsym}
\usepackage[T1]{fontenc}
\usepackage[utf8]{inputenc}
\usepackage{microtype}
\usepackage{inconsolata}
\usepackage{xspace}
\usepackage{graphicx}
\usepackage{amsmath}
\usepackage{newtxmath}
\usepackage{algorithm}
\usepackage{algpseudocode}

\newcommand{\algname}{\textsc{AttentionRAG}\xspace}

\NewDocumentCommand{\yixiong}{ mO{} }{\textcolor{green}{\textsuperscript{\textit{Yixiong}}\textsf{\textbf{\small[#1]}}}}
\NewDocumentCommand{\gu}{ mO{} }{\textcolor{blue}{\textsuperscript{\textit{Guxd}}\textsf{\textbf{\small[#1]}}}}
\NewDocumentCommand{\shi}{ mO{} }{\textcolor{orange}{\textsuperscript{\textit{Shiyl}}\textsf{\textbf{\small[#1]}}}}
\NewDocumentCommand{\tran}{ mO{} }{\textcolor{purple}{\textsuperscript{\textit{tsun}}\textsf{\textbf{\small[#1]}}}}

\newcommand{\secref}[1]{\S\ref{#1}}

\title{AttentionRAG: Attention-Guided Context Pruning in Retrieval-Augmented Generation}
\author{
  \textbf{Yixiong Fang}\textsuperscript{\rm 1}\thanks{~~Equal contribution.},
  \textbf{Tianran Sun}\textsuperscript{\rm 1}\footnotemark[1],
  \textbf{Yuling Shi}\textsuperscript{\rm 1},
  \textbf{Xiaodong Gu}\textsuperscript{\rm 1}\thanks{~~Corresponding author.} \\
  Shanghai Jiao Tong University \\
  \texttt{\{fangyixiong, Seriousss, yuling.shi, xiaodong.gu\}@sjtu.edu.cn}
}

\begin{document}

\maketitle

\renewcommand{\thefootnote}{\fnsymbol{footnote}}
\footnotetext[1]{Corresponding author.}
\renewcommand{\thefootnote}{\arabic{footnote}}

\maketitle
\begin{abstract}
    While RAG demonstrates remarkable capabilities in LLM applications, its effectiveness is hindered by the ever-increasing length of retrieved contexts, which introduces information redundancy and substantial computational overhead. Existing context pruning methods, such as LLMLingua, lack contextual awareness and offer limited flexibility in controlling compression rates, often resulting in either insufficient pruning or excessive information loss. In this paper, we propose \algname, an attention-guided context pruning method for RAG systems. The core idea of \algname lies in its attention focus mechanism, which reformulates RAG queries into a next-token prediction paradigm. This mechanism isolates the query’s semantic focus to a single token, enabling precise and efficient attention calculation between queries and retrieved contexts.
    Extensive experiments on LongBench and Babilong benchmarks show that \algname achieves up to 6.3x context compression while outperforming LLMLingua methods by around 10\% in key metrics.
\end{abstract}

\section{Introduction}

Retrieval-Augmented Generation (RAG)~\citep{lewis2021retrievalaugmentedgenerationknowledgeintensivenlp} demonstrated remarkable capabilities in Large Language Model (LLM) applications such as reasoning~\citep{huang2023reasoninglargelanguagemodels}, question-answering~\citep{rajpurkar2016squad100000questionsmachine,wang-etal-2024-leave}, and open-ended text generation~\citep{que2024hellobenchevaluatinglongtext}. Although LLMs have demonstrated superior performance, they often lack domain knowledge. By directly leveraging external documents, RAG provides a lightweight, cost-efficient solution to expand the LLMs' knowledge base without retraining their parameters. 

\begin{figure}[ht]
    \centering
    \includegraphics[width=\linewidth]{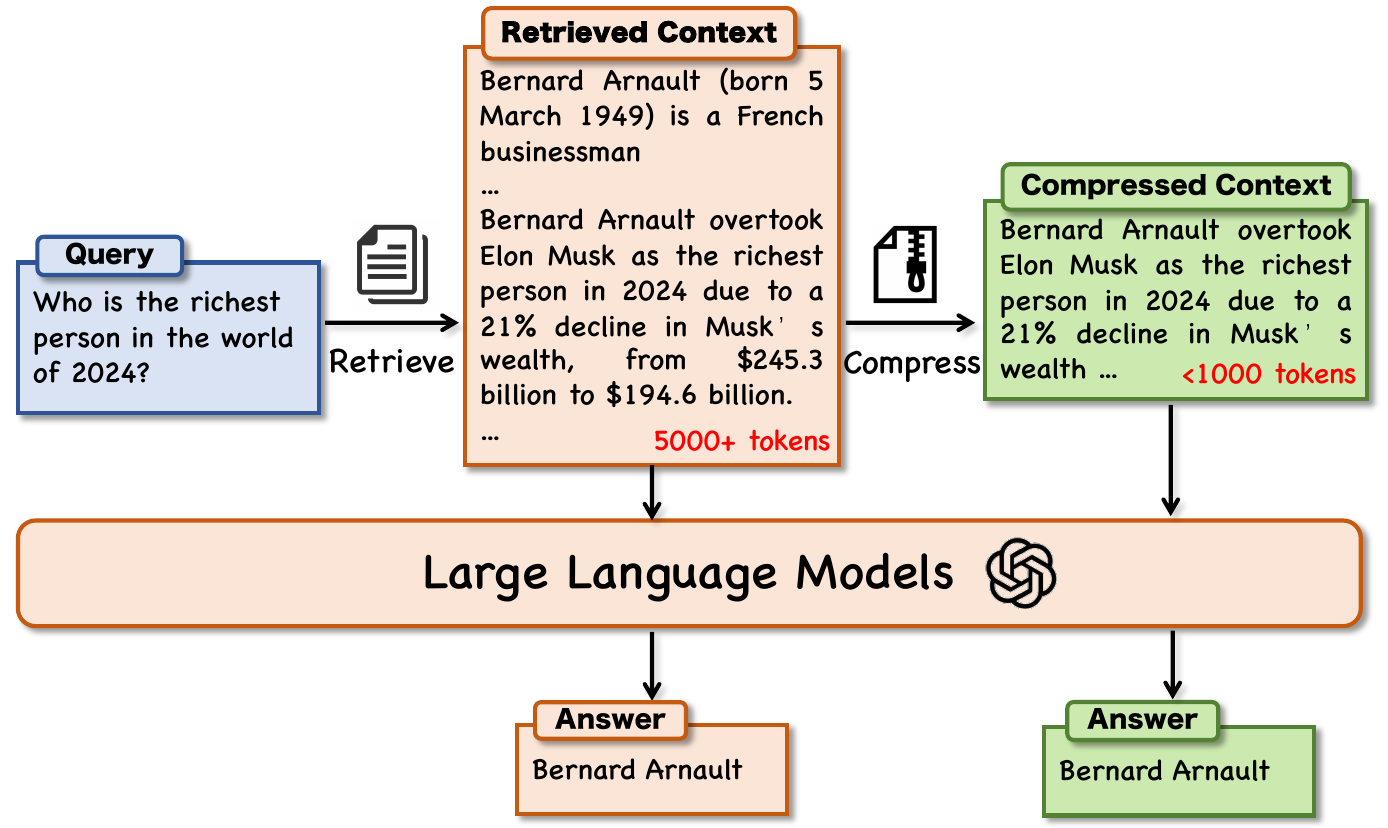}
    \caption{Illustration of RAG context compression}
    \label{fig:rag_compress}
\end{figure}

While effective, RAG-based systems encounter significant challenges in handling long contexts.
As the number of retrieved documents grows, the context becomes excessively long, introducing large amounts of redundant and irrelevant information. This issue, as highlighted by \citet{shi2024enhancingretrievalmanagingretrieval}, can lead to hallucinations and a decline in the performance of the LLM \citep{chiang-cholak-2022-overcoming}.

To mitigate these issues, recent work has explored context compression techniques~\citep{jiang2023llmlinguacompressingpromptsaccelerated,cheng2024xrag,verma2024survey}. 
For example, LLMLingua~\citep{jiang2023llmlinguacompressingpromptsaccelerated} compresses prompts by using a budget controller that allocates varying compression ratios to different components of the prompt, such as instructions and demonstrations. However, these methods lack context-awareness, making it challenging to determine the optimal compression ratio for a given LLM, resulting in context redundancy or over-compression.

\begin{figure*}[htbp]
    \centering
    \includegraphics[width=\linewidth, trim=80 115 100 80 clip]{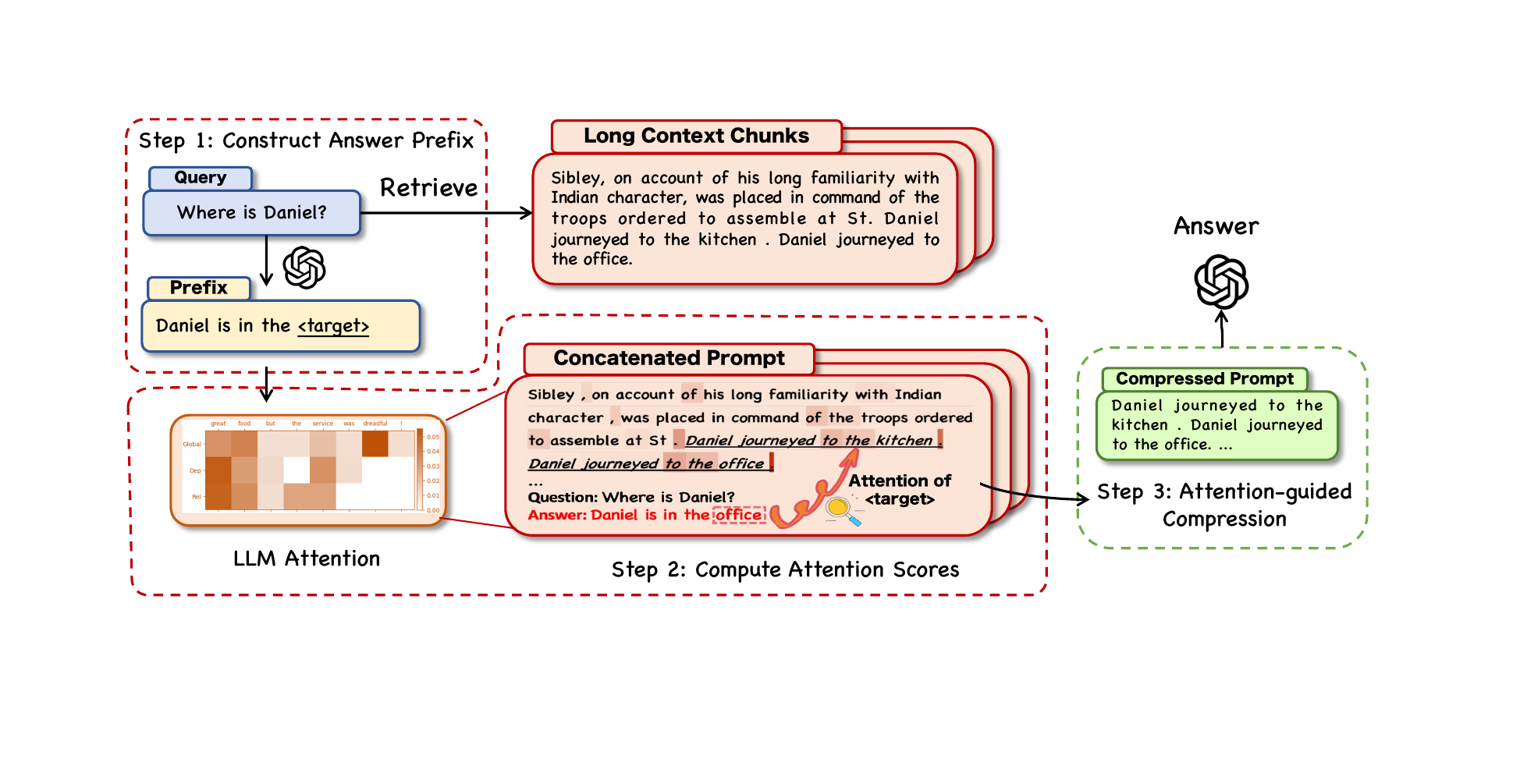}
    \caption{Illustration of \algname: 
1) We first generate a hint prefix using the original query; if generation fails, we fallback to a fixed hint prefix.  
2) The long context is then divided into smaller chunks for localized processing.  
3) For each chunk, we batch the input with the hint prefix (1 token only) and use an LLM to compute attention features over the full context, focusing on the generated anchor token.  
4) Using these attention features, we identify and extract the most relevant content from each chunk to construct a compressed context.
}
    \label{fig:structure}
\end{figure*}

Although attention mechanisms are central to LLMs for content selection, there is limited research on using them specifically for context pruning during retrieval. This gap is primarily due to two fundamental challenges that hinder the effectiveness of attention in the retrieval architecture:
(1) The long nature of RAG contexts exacerbates attention dilution~\citep{hsieh2024middlecalibratingpositionalattention,zhu2024acceleratinginferenceretrievalaugmentedgeneration}. When the context is excessively long, the attention scores are spread thin, causing the model to allocate less focus to any single token. This dilution of attention reduces the ability of the model to concentrate on the most relevant parts of the retrieved context;
and (2) The sentence-based nature of the query makes it challenging to directly align with critical content in the context.
The attention score distribution varies across tokens within a sentence, with some tokens focusing on semantic information while others attend to details~\citep{clark-etal-2019-bert,gu-etal-2022-continuous}. For instance, in the sequence ``Mary is in the car,'' the attention score of ``in'' tends to focus on semantic relationships, while ``car'' focuses more on a specific detail. 
Therefore, the query lacks the explicit or semi-structured reference points (such as keywords or pivotal terms) needed to guide the attention mechanism toward the most salient parts of the context. 

In this paper, we propose a novel methodology called \algname (Attention-Guided Retrieval-Augmented Generation), which improves the relevance of the information extracted from the retrieved context by leveraging attention scores from intermediate layers of LLMs. 
To enable attention across queries and context, \algname introduces an attention focus mechanism that isolates the query's semantic focus to a single token, enabling precise and efficient attention computation between queries and retrieved contexts in a single pass. 
More specifically, for each query (e.g, ``Where is Daniel?''), we construct an answer hint prefix (e.g., ``Daniel is in the \_\_\_\_'') in a <next-token-prediction> format. Next, we take the retrieved context, the query, and the constructed answer prefix as input to an LLM. The missing next token in the prefix stands for the focal point of the query, guiding the LLM's attention to each token in the context. Finally, we produce a compressed context by selecting sentences from the original context with the top-k attended tokens. 

Extensive experiments on LongBench~\citep{bai2024longbenchbilingualmultitaskbenchmark} and BABILong~\citep{kuratov2024babilongtestinglimitsllms} demonstrate that our method achieves up to 6.3x context compression while maintaining or exceeding the performance of uncompressed contexts. 
The results suggest that \algname not only facilitates the extraction of relevant information but also enhances the model's reasoning capabilities.
Particularly, it achieves these benefits without requiring additional training, making it highly adaptable across different models and practical for real-world applications. 

Our key contributions are as follows:
\begin{itemize}
\setlength{\itemsep}{0pt}
\setlength{\topsep}{0pt}
\setlength{\parsep}{0pt}
\item We propose a lightweight, transferable, and question-aware method for long-context pruning in RAG systems.
\item We introduce a novel attention focus mechanism by reformulating RAG queries into a next-token prediction template, enabling precise and efficient computations of attention between queries and retrieved contexts.
\item We conduct extensive experiments on LongBench and Babilong benchmarks. Results demonstrate the effectiveness of \algname in long-context RAG systems.
\end{itemize}

\section{Preliminaries}

\subsection{Retrieve-Augmented Generation}
Retrieve-Augmented Generation (RAG) is a framework that enhances the capabilities of LLMs by integrating external knowledge through retrieval. A RAG system typically consists of two components: a \textit{retriever}, which fetches relevant documents, called contexts, from a large corpus based on a query, and a \textit{generator}, which generates an answer using both the retrieved context and the model’s internal knowledge. This combination enables more accurate and contextually relevant outputs, especially for tasks requiring detailed or up-to-date information that might not be present in the model’s training data. 

\subsection{Attention Mechanism in LLMs}

The attention mechanism is a key component in modern LLMs, allowing the model to focus on different parts of the input sequence~\citep{vaswani2023attentionneed}. 
For a given context $c$, the mechanism calculates a score for each token $t\in c$ based on its relevance to other tokens: 
\begin{equation}
\text{Attention}_l(c,t) = \text{Softmax}\left(\frac{Q_l K_l^T}{\sqrt{d_k}}\right) V_l
\label{eq:attention_formula}
\end{equation}
where:
\( Q_l \) is the query matrix at layer \( l \),
\( K_l \) is the key matrix at layer \( l \),
\( V_l \) is the value matrix at layer \( l \),
\( d_k \) is the dimensionality of the key vectors.

The scores indicate how much ``focus'' the token receives from the model, providing insights into which tokens are most relevant in a given context.
This enables it to be used for text compression, where selecting tokens with high attention scores can reduce the input to the model while retaining the most important information~\citep{tarzanagh2023maxmargintokenselectionattention}. Although attention is central to LLMs, it suffers from dilution in handling long contexts \cite{hsieh2024middlecalibratingpositionalattention}.

\section{Method}\label{sec:Method}


In this work, we propose a novel approach to optimize RAG systems by compressing the retrieved context without compromising performance. 

\paragraph{Problem Formulation}
Given a query $q$, a RAG system retrieves a relevant set of documents \{$d_1,\ldots,d_K$\} from a text corpus $D$. The retrieved documents are concatenated into a retrieved context $c=d_1\oplus\ldots\oplus d_N$. A large language model $P_\phi(a|q,c)$ takes the question and the retrieved context $c$ as input and produces an answer $a$ to the question. 
Our goal is to compress the retrieved context $c$ into a dense one $c'$ such that $|c'|\ll|c|$ while the LLM maintains the quality of generated answers when taking $c'$ as input. 

In this paper, we propose a novel attention-guided context pruning method called \algname. The key idea of \algname is reformulating each RAG query into a next-token-prediction template (called answer hint prefix). This strategy allows the LLM to calculate the query-context attention through one token, therefore significantly improving the alignment between query and context, and reducing the time complexity for attention calculation. 

Figure \ref{fig:structure} shows the overall structure of our method. The pipeline involves three key steps:
First, we generate an answer hint prefix for each query (\secref{subsec:anchor_token}); Next, the generated prefix is appended to the original query and context as input to the LLM. The LLM is instructed to predict the follow-up token to the answer prefix, obtaining the attention scores (\secref{subsec:computing_attn}). 
Finally, we perform attention-guided compression: using attention scores from a language model, we identify and retain the most relevant parts of the retrieved context. 
Each of the steps is elaborated in the following sections.

\subsection{Construct Answer Hint Prefix}\label{subsec:anchor_token}
To improve the alignment between the query and context, we associate each query with an answer hint prefix that allows the LLM to calculate the query-context attention through one focal token. 

For each query, an \textit{answer hint prefix} is defined as an incomplete answer to the query in a \textit{next-token-prediction} format, where the blank token to be predicted serves as the focal token of the query, directing the LLM's attention to the most relevant parts of the context. For instance, as shown in Figure~\ref{fig:structure}, for the query ``Where is Daniel?'', the corresponding answer hint prefix can be ``Daniel is in the \_\_\_''. When we take the context, query, and the answer hint prefix as input to the LLM, the token to be predicted by the model, such as ``park'', becomes the focal point of the query, directing the model's attention to the most relevant parts of the context and ensuring that the attention calculation focuses on the crucial information related to the query.

We prompt the LLM to construct the answer hint prefix\footnote{We use GPT-4o Mini for the generation, the prompt detail and fixed hint prefix can be referred to ~\secref{sec:app_hint} and ~\secref{sec:app_fixed}.}.
In detail, according to the query's grammatical attributes, we categorize the answer hint prefix into two types: empty and non-empty ones.
For example, for queries like wh-questions, the hint prefix can be derived from the query itself.
In contrast, queries like yes/no-questions, where the answer's first token is "Yes/No", already align with the next-token-prediction paradigm. 
However, some type of questions may not have such hint prefix (like summarization task), we also provide a fixed hint prefix that prompt the model to output the keyword of context, ensuring robustness of our method.
Our fixed hint prefix is: \textit{"Please output the most relevant keyword or phrase that is relevant to the answer of the question."}
Leveraging the semantic understanding capabilities of LLMs, we prompt them with example answer hint prefix of various types questions to automatically determine the type and generate the answer prefix hint.

By focusing attention on the focal token, this approach enhances both the precision and efficiency of the attention mechanism. The single-token focus accelerates computations by reducing the number of tokens involved in attention calculations. Simultaneously, concentrating on a target token—such as "car" in the sentence "Daniel is in the car"—enables the model to effectively identify and prioritize the most relevant information in the context. 


\subsection{Compute Attention Features from LLM}\label{subsec:computing_attn}

In this part, we aim to compute attention features based on the focal token. To address the issue of diluted attention scores, we divide the retrieved context \( C \) produced by the RAG framework into smaller chunks. We adopt a uniform chunking strategy, assuming each chunk consists of \( m \) tokens. Let \( c_1, \dots, c_n \) represent the resulting chunks, where \( n = \lceil |C| / m \rceil \). For each chunk \( c_j \), we concatenate the chunked context, query, and instruction (we instruct the LLM to generate "none" after the hint prefix if the chunk is irrelevant, which will be used in \secref{subsec:attention_compression}) and the generated answer hint prefix and feed this into the LLM. The LLM is then instructed to perform next-token prediction and compute the attention scores.

We define $a_j$ as the first token generated following the answer hint prefix in chunk $j$. The attention feature $A_j$ for $a_j$ is computed as the sum of attention scores over all layers of the model, focusing on context tokens:
\begin{equation}
A_j = \sum_{l=0}^{L} \text{Attention}_l(c_j, a_j)
\label{eq:attention_feature}
\end{equation}
where $L$ is the total number of layers in the model, and $\text{Attention}_l(c_j, a_1)$ is the attention score at layer $l$ for the token $a_1$ relating to the context chunk $c_j$. This score is computed by the self-attention mechanism at each layer, capturing both local and global dependencies in the input.


The total attention feature $A_i$ reflects the model's focus on the most relevant components of the input when generating the first token, and is the sum of the attention values across all layers. We choose to sum across all layers for analysis because the attention distribution in each layer can vary depending on the task. For easier tasks, earlier layers may already capture sufficient information to generate the final answer, while for more difficult tasks, the model might rely on the later layers~\citep{jin2025exploringconceptdepthlarge}. Since the function of each layer can vary from task to task, focusing on a single layer or a subset of layers could introduce bias in the attended information. To mitigate this issue, we sum the attention across all layers, which helps reduce task-specific bias. The choice of attention layers will be further explored in the ablation study in ~\secref{subsec:ablation}.

\subsection{Compress with Attention}\label{subsec:attention_compression}
For each chunk, after generating the focal token, we first check whether this token is "none." If this is the case, we skip the chunk, as it is deemed irrelevant to the task. If the focal token is valid, we proceed by identifying the tokens in the context that have the highest attention features with respect to the focal token. These attention features represent how much each token in the context is relevant to the focal token, which serves as the focal point of the query.

Next, we select the top-\( k \) tokens based on their attention features, as these tokens are considered the most relevant to the focal token. To ensure that the context used for generating the final response is both relevant and concise, we focus on the sentences that contain these top-\( k \) tokens. By selecting these sentences, we retain the information most pertinent to the focal token. These selected sentences are then concatenated to form a compressed context \( c_j^\prime \). 
\begin{equation}
\resizebox{\linewidth}{!}{$
c_j^\prime = \text{Concat}\left(\left\{ s \mid t_r\in\text{Top-}k(A_j) \text{ and } t_r\in s \right\} \right)
$}
\label{eq:compression_formula}
\end{equation}
where $s$ denotes a selected sentence.

\subsection{Time Efficiency and Batch Generation}
Since we employ a next-token prediction paradigm, only one focal token needs to be generated for each chunk. Furthermore, as each chunk is processed independently, batch generation can be used to accelerate the process. This approach results in high time efficiency. Moreover, we can use quantified model to further accelerate the compression process.
We provide the pseudocode of our method in Algorithm 1 in Appendix.

\section{Experimental Setup}
In this experiment, we evaluate the efficacy of \algname in long context compression. 

\subsection{Datasets}
Due to fluctuations in the experimental results, each experiment was conducted three times, and the final result is the average of these trials.
\paragraph{Long Context Reasoning} For our experiments, we incorporate datasets TriviaQA~\citep{joshi2017triviaqa}, HotpotQA~\citep{yang2018hotpotqa}, and 2WikiMQA~\citep{ho2020constructing} from Longbench~\citep{bai2024longbenchbilingualmultitaskbenchmark}: TriviaQA assesses fact retrieval over long contexts, HotpotQA emphasizes multi-hop reasoning across dispersed clues, and 2WikiMQA tests the model's ability to synthesize information from multiple sources. 
\paragraph{Babilong} \citep{kuratov2024babilongtestinglimitsllms}: 
The BABILong benchmark provides a comprehensive framework for evaluating an LLM's ability to reason and retrieve over long contexts. To assess \algname's capabilities on relatively shorter and sparse content, we select test splits of 1kqa1, 2kqa1, and 4kqa1 tokens to demonstrate its performance across different circumstances.
\paragraph{Long Context Summarization} To demonstrate the robustness of \algname, we also evaluate it on GovReport~\citep{huang-etal-2021-efficient} from Longbench. In this task, we employ fixed hint prefix, showing the stability of our method.

\subsection{Metrics} 
We measure the effectiveness of compression using three metrics: 

\noindent\textbf{Exact Match (EM)}: Measures the percentage of predicted answers that exactly match the ground-truth answers.

\sloppy
\noindent\textbf{LLM-as-a-judge scores}: We leverage GPT-4~\cite{openai2024gpt4technicalreport} to assess the correctness of the model-generated answers. Specifically, we input the question, the model-generated answer, and the ground truth answer into GPT-4o, asking it to determine whether the provided answer is correct. 
The prompt used for the scoring can be found in ~\secref{sec:appendix}.

\noindent\textbf{Compression Ratio (CR)}: The ratio of tokens in the original context to the compressed context, defined as {\small \( CR = \frac{\text{\# tokens in original context}}{\text{\# tokens in compressed context}} \)}.


\subsection{Baselines}
We compare \algname with state-of-the-art context compression baselines, including LLMLingua2~\citep{pan2024llmlingua2datadistillationefficient} and LongLLMLingua~\citep{jiang2024longllmlinguaacceleratingenhancingllms}, which represent question-unaware and question-aware approaches, respectively. This diverse selection underscores the robustness of \algname. For fair comparison, we match or exceed their compression rates. We also include the uncompressed context (``Original Prompt''), a random compression baseline (``Random''), and a vanilla RAG baseline using BGE-M3~\citep{chen2024bgem3embeddingmultilingualmultifunctionality}.

\subsection{LLMs for Compression}
We select two open-source LLMs for evaluation, Llama-3.1-8B-Instruct~\citep{grattafiori2024llama3herdmodels} and Qwen-2.5-7B-Instruct~\citep{qwen2025qwen25technicalreport}. 
We experiment with both 8B and 70B versions of Llama-3.1-Instruct as our compression model, denoted as \algname(8B) and \algname(70B) respectively. Detailed hyperparameter configurations are provided in the Appendix~\secref{subsec:hyperparameter_settings}.

\section{Results}

\subsection{Overall Results}

\begin{table*}[t]
    \centering
    \setlength{\tabcolsep}{4pt}
    \renewcommand{\arraystretch}{1}
    \resizebox{\linewidth}{!}{
    \begin{tabular}{lcccccccccccc}
    \toprule
    \multirow{2}{*}{Model} & \multirow{2}{*}{Method} & \multicolumn{3}{c}{2WikiMQA} & \multicolumn{3}{c}{HotpotQA} & \multicolumn{3}{c}{TriviaQA} & \multicolumn{2}{c}{Gov} \\
    \cmidrule(lr){3-5}\cmidrule(lr){6-8}\cmidrule(lr){9-11}\cmidrule(lr){12-13}
      & & EM$\uparrow$ & LLM Judge$\uparrow$ & CR$\uparrow$
        & EM$\uparrow$ & LLM Judge$\uparrow$ & CR$\uparrow$
        & EM$\uparrow$ & LLM Judge$\uparrow$ & CR$\uparrow$
        & Rouge$\uparrow$ & CR$\uparrow$ \\
    \midrule\midrule
    \multirow{9}{*}{Llama-3.1-8B-Instruct}
        & Original Prompt     & 0.49 & 0.46 & 1.0x & 0.50 & 0.62 & 1.0x & 0.87 & 0.80 & 1.0x & 0.32 & 1.0x\\
        & Random              & 0.21 & 0.15 & 5.0x & 0.05 & 0.01 & 3.3x & 0.76 & 0.73 & 1.7x & --   & --  \\
        & LLMLingua2 (small)  & 0.24 & 0.26 & 5.0x & 0.40 & 0.45 & 3.3x & 0.86 & \textbf{0.82} & 1.7x & 0.23 & 3.3x\\
        & LLMLingua2 (large)  & 0.33 & 0.30 & 5.0x & 0.44 & 0.51 & 3.3x & 0.84 & 0.76 & 1.7x & 0.24 & 3.3x\\
        & LongLLMLingua       & 0.23 & 0.22 & 3.3x & 0.33 & 0.40 & 3.3x & 0.83 & 0.72 & 1.7x & 0.23 & 3.1x\\
        & BGE-M3              & 0.40 & 0.35 & 3.5x & 0.42 & 0.50 & 3.0x & 0.80 & 0.77 & 1.9x & 0.21 & 3.0x\\
        & \algname(small)     & 0.39 & 0.36 & 6.3x & 0.45 & 0.55 & 3.7x & 0.84 & 0.77 & 2.0x & \textbf{0.28} & 3.4x\\
        & \algname(large)     & \textbf{0.42} & \textbf{0.38} & 15x & \textbf{0.48} & \textbf{0.61} & 5.6x & \textbf{0.89} & \underline{0.81} & 1.7x & \textbf{0.28} & 4.2x\\
    \midrule\midrule
    \multirow{9}{*}{Qwen2.5-7B-Instruct}
        & Original Prompt     & 0.56 & 0.46 & 1.0x & 0.58 & 0.68 & 1.0x & 0.94 & 0.82 & 1.0x & 0.31& 1.0x\\
        & Random              & 0.18 & 0.06 & 5.0x & 0.04 & 0.01 & 3.3x & 0.65 & 0.51 & 1.7x & --   & --  \\
        & LLMLingua2 (small)  & 0.29 & 0.27 & 5.0x & 0.43 & 0.47 & 3.3x & \textbf{0.91} & 0.80 & 1.7x & 0.21 & 3.3x\\
        & LLMLingua2 (large)  & 0.40 & 0.28 & 5.0x & 0.48 & \textbf{0.58} & 3.3x & 0.87 & \textbf{0.81} & 1.7x & 0.22 & 3.3x\\
        & LongLLMLingua       & 0.31 & 0.18 & 4.0x & 0.28 & 0.31 & 3.3x & 0.83 & 0.73 & 1.7x & 0.23 & 3.1x\\
        & BGE-M3              & 0.40 & 0.35 & 3.5x & 0.42 & 0.50 & 3.0x & 0.80 & 0.77 & 1.9x & 0.22& 3.0x\\
        & \algname(small)     & \textbf{0.41} & 0.28 & 6.3x & 0.51 & 0.54 & 3.7x & 0.83 & 0.73 & 2.0x & \textbf{0.28} & 3.4x\\
        & \algname(large)     & \textbf{0.41} & \textbf{0.30} & 15x & \textbf{0.53} & \underline{0.56} & 5.6x & \underline{0.88} & \underline{0.80} & 1.7x & \textbf{0.28} & 4.2x\\
    \bottomrule
    \end{tabular}}
    \caption{Performance comparison on QA datasets (2WikiMQA, HotpotQA, TriviaQA) and the Gov summarization dataset.}
    \label{table:experiment_results}
    \vspace{-0.15in}
\end{table*}

\begin{table*}[ht]
    \centering
    \setlength{\tabcolsep}{4pt}
    \renewcommand{\arraystretch}{1}
    \resizebox{\linewidth}{!}{
    
    \begin{tabular}{lcccccccccc}
    \toprule
    \multirow{2}{*}{Model} & \multirow{2}{*}{Method} & \multicolumn{3}{c}{1K} & \multicolumn{3}{c}{2K} & \multicolumn{3}{c}{4K} \\ 
    \cmidrule(lr){3-5} \cmidrule(lr){6-8} \cmidrule(lr){9-11}
    & & EM$\uparrow$ & LLM Judge$\uparrow$ & CR$\uparrow$ & EM$\uparrow$ & LLM Judge$\uparrow$ & CR$\uparrow$ & EM$\uparrow$ & LLM Judge$\uparrow$ & CR$\uparrow$ \\
    \midrule
    \midrule
    \multirow{7}{*}{Llama-3.1-8B-Instruct} 
        & Original Prompt & 0.74& 0.64& 1.0x& 0.67& 0.57& 1.0x & 0.71& 0.60&1.0x\\
    \midrule
        & Random &  0.11&  0.04&  2.9x&   0.09&    0.05&    3.3x&     0.05&     0.04&   3.3x\\
        & LLMLingua2 (small) & 0.55 & 0.41 & 2.9x & 0.38 & 0.29 & 3.3x & 0.36& 0.29 & 3.3x\\
        & LLMLingua2 (large) & 0.69 & 0.51 & 2.9x & 0.53 & 0.41 & 3.3x & 0.49 & 0.37& 3.3x\\
        & LongLLMLingua & 0.63 & 0.59& 2.9x &0.43& 0.38 & 3.3x & 0.47 & 0.36& 3.3x\\
        & \algname(small) &\textbf{0.74}& \textbf{0.65} & 3.8x & \textbf{0.58}& \textbf{0.49}& 3.8x &\textbf{0.53}&  \textbf{0.41}& 5.6x \\
    \midrule
    \midrule
    \multirow{7}{*}{Qwen2.5-7B-Instruct} 
        & Original Prompt & 0.87& 0.78& 1.0x & 0.75& 0.71& 1.0x &0.80& 0.75& 1.0x \\
    \midrule
        & Random &  0.04&  0.04&  2.9x&   0.05&    0.03&    3.3x&     0.04&     0.01&   3.3x\\
        & LLMLingua2 (small) & 0.49 & 0.38 &2.9x & 0.37& 0.30 & 3.3x & 0.34 & 0.24 & 3.3x \\
        & LLMLingua2 (large) & 0.70 & 0.52 & 2.9x & 0.55 & 0.41 & 3.3x&0.50 &0.38 & 3.3x\\
        & LongLLMLingua & 0.54 & 0.47 & 2.9x & 0.35 & 0.29& 3.3x & 0.37 & 0.29 & 3.3x \\
        & \algname(small) & \textbf{0.88} & \textbf{0.76}& 3.8x & \textbf{0.62}& \textbf{0.57}& 3.8x & \textbf{0.66}& \textbf{0.59}& 5.6x \\
    \bottomrule
    \end{tabular}
    }
    \caption{Comparison of different methods on BABILong.}
    \label{table:experiment_results_babilong}
\end{table*}
    
As the results in Tables \ref{table:experiment_results} and \ref{table:experiment_results_babilong} show,
\algname outperforms nearly all baseline methods across the benchmarks, consistently achieving superior results. 
On the LongBench benchmark, which includes contexts of tens of thousands of tokens, \algname outperforms the LLMLingua methods in most metrics, particularly in BABILong 1k, where it achieves an 18\% higher score than the best LLMLingua model. 
Specifically, in TriviaQA, \algname outperforms the uncompressed method by 2\% in EM score, demonstrating the effectiveness of our approach. However, the low compression ratio is due to two parallel factors: the scattering of relevant information across the context and the use of uniform hyperparameters.
All these factors contribute to the lower ratio, which we aim to enhance. We discuss this in more detail in the Ablation Study.
Also, our method excels in summarization tasks, showing the stability of our method.
Similar results can be observed in the BABILong benchmark, which involves a wider range of RAG context lengths. Our method consistently outperforms all LLMLingua methods while achieving the highest compression ratio, further demonstrating the effectiveness of \algname in varying context sizes.
\vspace{-5pt}
\subsection{Efficiency Analysis}\label{sec:efficiency}
As discussed in~\secref{sec:Method}, we divide the context into smaller chunks and use attention mechanisms for compression. The overall time cost is spent in the forward pass of the compression model plus the answering of the generation model. We employ LLaMA-3.1-Instruct with int4 quantization for compression. We choose HotpotQA, where the average context length is about tens of thousands tokens. As shown in Table~\ref{tab:hotpotqa_inference}, the overall time cost is significantly reduced compared to baseline methods, and the quantized model can also ensure similar performance on results. 
\begin{table*}[t]
\centering
\renewcommand{\arraystretch}{1}
\begin{tabular}{lccccc}
\toprule 
\multirow{2}{*}{\bf Method} & \multirow{2}{*}{\bf EM $\uparrow$} & \multirow{2}{*}{\bf LLM Judge $\uparrow$} & \multirow{2}{*}{\bf CR $\uparrow$ } & \multicolumn{2}{c}{\bf Latency} \\
\cline{5-6}
 &   &   &  & \bf Compression & \bf Answering \\
\midrule
Original Prompt & 0.50 & 0.62 & 1.0x & -- & 19.36 \\
Vanilla RAG (BGE-M3) & 0.42 & 0.50 & 3.0x & 6.11 & 5.99 \\
LLMLingua2 (small) & 0.40 & 0.45 & 3.3x & 0.44 & 5.39 \\
LLMLingua2 (large) & 0.44 & 0.51 & 3.3x & 0.88 & 5.43 \\
LongLLMLingua & 0.42 & 0.52 & 3.3x & 7.12 & 5.41 \\
\textbf{AttentionRAG} & \textbf{0.45} & \textbf{0.54} & \textbf{4.0x} & \textbf{3.99} & \textbf{4.94} \\
\bottomrule
\end{tabular}
\caption{Time latency for inference by various methods on HotpotQA}
\label{tab:hotpotqa_inference}
\end{table*}

\subsection{Case Study}
Table~\ref{tab:case_study} presents a practical example of \algname, demonstrating how our method effectively compresses the context while maintaining both accuracy and readability. Unlike the original context, which contains redundant information that can negatively affect the response quality, \algname generates a concise and coherent output with minimal token usage. Other methods, such as LLMLingua2, while providing a more compact result, produce fragmented and less readable responses that lose coherence and relevance. Similarly, LongLLMLingua, despite reducing the context, fails to provide a clear and focused answer. In contrast, \algname generates the correct answer, “Ozalj,” with the highest compression ratio, illustrating its ability to preserve the essential information. This highlights \algname’s capacity to enhance overall response quality, effectively balancing compression and clarity without introducing unnecessary complexity.
\begin{table*}[!t]

\centering
\small
\renewcommand{\arraystretch}{1.1} 
\begin{tabular}{lp{10cm}p{2cm}}
\toprule
\textbf{Query}: & Where was the wife of Francis I Rákóczi born? & \bf Answer: Ozalj \\ \hline
\textbf{Original Context}: & Passage 1:
Waldrada of Lotharingia
Waldrada was the mistress, and later the wife, of Lothair II of Lotharingia.
Biography
Waldrada's family origin is uncertain. The prolific 19th-century French writer Baron Ernouf suggested that Waldrada was of noble Gallo-Roman descent, sister of Thietgaud... (\textbf{7003 tokens}) 
& city of Gyulafehérvár, Transylvania. \textcolor{red}{$\times$}
 \\ 
 \midrule
\textbf{Random}: & drada,, of.
th-century French Baron Er of-R sister ofga bishopther arch of, not any socialoli,... (\textbf{1400 tokens}) &  The text does mention it. \textcolor{red}{$\times$}\\ 
\textbf{LLMLingua2}: & Waldrada Lotharingia mistress Lothair II Gallo sister Thietgaud niece Gunther Vita Sancti related Eberhard II Etichonids 855 Lothar II married Teutberga 858 862 Nicholas 863Charles ...(\textbf{632 tokens})& Munkács.\textcolor{red}{$\times$} \\ 
\textbf{LongLLMLingua}: & Passage:Waldrada theressairia. is The proific 1th French Baron Ern thatadaoman, sister of Th Trier, Gun of and have suggested of social though anatic.itactoli thatada Ehard II,edbourgichon . ... (\textbf{920 tokens}) & Hungary. \textcolor{red}{$\times$} \\ 
\midrule
\bf \algname: & ... Life Early years and family \colorbox{green!30}{Ilona was born Ilona Zrínyi in Ozalj} ... She was the daughter of Petar Zrinski, Ban (viceroy) of Croatia, the niece of both Miklós Zrínyi and Fran Krsto Frankopan and \colorbox{green!30}{the wife of Francis Rákóczi I} ... (\textbf{273 tokens})       
 & Ozalj \textcolor{green!80}{$\checkmark$}\\
 \bottomrule
\end{tabular}
\caption{Examples of compression results by various methods} 
\label{tab:case_study}
\end{table*}

\subsection{Ablation Study}\label{subsec:ablation}

\noindent\textbf{Fixed Hint Prefix}
As discussed in~\secref{subsec:anchor_token}, we use a fixed hint prefix for questions that cannot generate one dynamically. The summarization performance under this setting is reported in Table~\ref{table:experiment_results}. To further validate the robustness of this approach, we conduct experiments on three additional datasets: 2WikiMQA, HotpotQA, and TriviaQA. As shown in Table~\ref{tab:fixed_hint}, we experiment with LLaMA-3.1-8B-Instruct, using only a fixed hint prefix results in a slight performance drop compared to our original method, yet still consistently outperforms all baselines. This demonstrates that even a static hint prefix serves as an effective anchor token, reliably guiding the model to identify important information within the context.

\begin{table}[t]
\centering
\renewcommand{\arraystretch}{1}
\resizebox{0.8\linewidth}{!}{
\begin{tabular}{lccc}
\toprule
Dataset & EM$\uparrow$ & LLM Judge$\uparrow$ & CR$\uparrow$ \\
\midrule
HotpotQA   & 0.44 & 0.53 & 3.3x \\
2WikiMQA   & 0.44 & 0.37 & 5.2x \\
TriviaQA   & 0.81 & 0.75 & 2.0x \\
\bottomrule
\end{tabular}
}
\caption{Performance of \algname using fixed hint prefix across datasets.}
\label{tab:fixed_hint}
\end{table}
\noindent\textbf{Combination with other RAG methods} To explore the integration potential of our method with retrieval-based techniques, we implemented a two-stage pipeline: we first use vanilla RAG (BGE-M3) for initial context retrieval, followed by our method for compression. Specifically, we retain the RAG chunk size at 300 tokens, while retrieve the top 15 relevant chunks, and then apply our compression method. We use LLaMA-3.1-8B-Instruct as the generation model and conduct experiments on the HotpotQA, 2WikiMQA, and TriviaQA benchmarks (See Table~\ref{tab:combination}).
\begin{table}[t]
\centering
\renewcommand{\arraystretch}{1}
\resizebox{0.8\linewidth}{!}{
\begin{tabular}{lccc}
\toprule
EM&	LLM &Judge&	CR\\
HotpotQA&	0.46&0.58	&9.1x (+5.4x)\\
2wikimqa&	0.43	&0.40&	7.9x (+1.6x)\\
triviaqa	&0.88	&0.83&	3.2x (+1.5x)\\
\bottomrule
\end{tabular}
}
\caption{Combination with other RAG method}
\label{tab:combination}
\end{table}




\noindent\textbf{Hyperparameters}
We conduct ablation studies on two hyperparameters: chunk size (\secref{sec:Method}) and the number of Top-$K$ tokens used to select sentences within each chunk (\secref{subsec:attention_compression}). For long contexts (e.g., LongBench), we use larger chunks and higher $K$ to handle dispersed information; for shorter contexts (e.g., Babilong), we use smaller chunks and lower $K$ for finer granularity.\footnote{Details in \secref{sec:appendix}.}
Using TriviaQA, we test various configurations and observe (Table~\ref{tab:chunk_k_ablation}) that increasing chunk size and reducing $K$ lowers the compression ratio with minimal performance drop. This suggests that dynamic hyperparameter tuning offers a better trade-off between efficiency and accuracy than fixed-ratio compression.

\begin{table}[t]
\centering
\renewcommand{\arraystretch}{1}
\resizebox{0.95\linewidth}{!}{
\begin{tabular}{ccccl}
\toprule
Chunk Size& Top-K & EM$\uparrow$ & LLM Judge $\uparrow$ &CR$\uparrow$ \\
\midrule
300 & 5  & 0.80 & 0.73  &\textbf{3.2x}\\
300 & 10 & 0.84 & 0.75  &2.2x\\
300 & 15 & \textbf{0.87} & \textbf{0.77}  &1.9x\\
\midrule
100 & 10 &\textbf{0.90} & \textbf{0.79} &1.9x\\
200 & 10 & 0.88 & 0.77  &2.2x\\ 
400 & 10 & 0.85 & 0.77  &\textbf{2.4x}\\
\bottomrule
\end{tabular}
}
\caption{Performance on TriviaQA with different chunk sizes and top-K values}
\label{tab:chunk_k_ablation}
\end{table}

\noindent\textbf{Size of Compression Models}
To evaluate the model-size sensitivity of our compression approach, we compare the performance of \algname when using LLaMA-3.1-Instruct 8B versus 70B to compute attention scores. As shown in Table~\ref{table:experiment_results}, the performance gap is minimal, indicating that even a lightweight model can effectively capture the attention patterns needed for compression. This suggests that \algname is largely agnostic to model size and can maintain strong performance without relying on large-scale models—offering substantial efficiency benefits. Furthermore, as demonstrated in \secref{sec:efficiency}, our method can be further accelerated through quantization. Overall, \algname is highly scalable and future-proof, with the potential to continuously benefit from advances in foundation model development while remaining efficient and adaptable.
\section{Related Work}
\paragraph{Retrieval-Augmented Generation}
RAG has shown strong performance in tasks like open-domain QA~\citep{han2024ragqaarenaevaluatingdomain,liu2024rag}, visual QA~\citep{wang2025vidorag,wang2025vrag}, and many other tasks~\citep{qiu2025latent}. But its effectiveness is often hindered by noisy or redundant retrieved content~\citep{shi2023largelanguagemodelseasily}. To address this, recent work has focused on improving retrieval quality. \citet{wang2023learningfiltercontextretrievalaugmented} trains models to filter irrelevant content, while \citet{xu2023recompimprovingretrievalaugmentedlms} uses extraction-based compression to retain key information.
Unlike these approaches, \algname does not require additional training. It leverages internal attention signals to identify informative content, offering strong performance and broad transferability across tasks and models.
\vspace{-5pt}
\paragraph{Prompt Compression}
To reduce the cost of long-context generation, both soft and hard prompt compression methods have been proposed. Soft prompt methods include Gist~\citep{mu2024learningcompresspromptsgist} and 500x Compressor~\citep{li2024500xcompressorgeneralizedpromptcompression}, which compress context into dense tokens with minimal loss. Hard prompt approaches like LLMLingua~\citep{jiang2023llmlinguacompressingpromptsaccelerated}, LongLLMLingua~\citep{jiang2024longllmlinguaacceleratingenhancingllms}, and LLMLingua2~\citep{pan2024llmlingua2datadistillationefficient} use token-level filtering, achieving substantial speedups and compression.
In contrast, \algname enhances LLM performance by selecting explainable, model-attended content without retraining. We compare against LongLLMLingua and LLMLingua2 to demonstrate its efficiency and robustness.

\vspace{-10pt}
\section{Conclusion}
\vspace{-10pt}
In this paper, we propose \algname, a novel attention-guided context pruning method for RAG systems. The core part of our method is the formatted attention focus mechanism, which constructs an answer hint prefix and utilize a fixed hint prefix in a \textit{next-token-prediction} format for each query, guiding the LLM to attend relevant tokens in the retrieved context through one token. We conduct extensive experiments on the 2WikiMQA, HotpotQA, TriviaQA, GovReport and Babilong benchmarks, demonstrating its strong performance.

\section*{Limitation}
In this section, we faithfully discuss the current limitations and potential avenues for future research.


Regarding the attention feature computation, we currently aggregate attention scores across all layers. However, we believe this process can be optimized using more sophisticated algorithms to improve efficiency.

Additionally, while we propose a dynamic compression ratio, we have not yet developed methods for explicitly controlling or instructing the desired ratio. Determining and setting precise parameters to achieve a specific compression ratio is a challenging task. In future work, we aim to investigate ways to provide more flexible and accurate control over the compression ratio.
\bibliography{custom}

\begin{thebibliography}{40}
\providecommand{\natexlab}[1]{#1}

\bibitem[{Bai et~al.(2024)Bai, Lv, Zhang, Lyu, Tang, Huang, Du, Liu, Zeng, Hou, Dong, Tang, and Li}]{bai2024longbenchbilingualmultitaskbenchmark}
Yushi Bai, Xin Lv, Jiajie Zhang, Hongchang Lyu, Jiankai Tang, Zhidian Huang, Zhengxiao Du, Xiao Liu, Aohan Zeng, Lei Hou, Yuxiao Dong, Jie Tang, and Juanzi Li. 2024.
\newblock \href {https://arxiv.org/abs/2308.14508} {Longbench: A bilingual, multitask benchmark for long context understanding}.
\newblock \emph{Preprint}, arXiv:2308.14508.

\bibitem[{Ben-Artzy and Schwartz(2024)}]{benartzy2024attendfirstconsolidatelater}
Amit Ben-Artzy and Roy Schwartz. 2024.
\newblock \href {https://arxiv.org/abs/2409.03621} {Attend first, consolidate later: On the importance of attention in different llm layers}.
\newblock \emph{Preprint}, arXiv:2409.03621.

\bibitem[{Chen et~al.(2024)Chen, Xiao, Zhang, Luo, Lian, and Liu}]{chen2024bgem3embeddingmultilingualmultifunctionality}
Jianlv Chen, Shitao Xiao, Peitian Zhang, Kun Luo, Defu Lian, and Zheng Liu. 2024.
\newblock \href {https://arxiv.org/abs/2402.03216} {Bge m3-embedding: Multi-lingual, multi-functionality, multi-granularity text embeddings through self-knowledge distillation}.
\newblock \emph{Preprint}, arXiv:2402.03216.

\bibitem[{Cheng et~al.(2024)Cheng, Wang, Zhang, Ge, Chen, Wei, Zhang, and Zhao}]{cheng2024xrag}
Xin Cheng, Xun Wang, Xingxing Zhang, Tao Ge, Si-Qing Chen, Furu Wei, Huishuai Zhang, and Dongyan Zhao. 2024.
\newblock xrag: Extreme context compression for retrieval-augmented generation with one token.
\newblock \emph{arXiv preprint arXiv:2405.13792}.

\bibitem[{Chiang and Cholak(2022)}]{chiang-cholak-2022-overcoming}
David Chiang and Peter Cholak. 2022.
\newblock \href {https://doi.org/10.18653/v1/2022.acl-long.527} {Overcoming a theoretical limitation of self-attention}.
\newblock In \emph{Proceedings of the 60th Annual Meeting of the Association for Computational Linguistics (Volume 1: Long Papers)}, pages 7654--7664, Dublin, Ireland. Association for Computational Linguistics.

\bibitem[{Clark et~al.(2019)Clark, Khandelwal, Levy, and Manning}]{clark-etal-2019-bert}
Kevin Clark, Urvashi Khandelwal, Omer Levy, and Christopher~D. Manning. 2019.
\newblock \href {https://doi.org/10.18653/v1/W19-4828} {What does {BERT} look at? an analysis of {BERT}`s attention}.
\newblock In \emph{Proceedings of the 2019 ACL Workshop BlackboxNLP: Analyzing and Interpreting Neural Networks for NLP}, pages 276--286, Florence, Italy. Association for Computational Linguistics.

\bibitem[{Dubey et~al.(2024)Dubey, Jauhri, Pandey, Kadian, Al-Dahle, Letman, Mathur, Schelten, Yang, Fan et~al.}]{grattafiori2024llama3herdmodels}
Abhimanyu Dubey, Abhinav Jauhri, Abhinav Pandey, Abhishek Kadian, Ahmad Al-Dahle, Aiesha Letman, Akhil Mathur, Alan Schelten, Amy Yang, Angela Fan, et~al. 2024.
\newblock \href {https://arxiv.org/abs/2407.21783} {The llama 3 herd of models}.
\newblock \emph{Preprint}, arXiv:2407.21783.

\bibitem[{et~al.(2024)}]{openai2024gpt4technicalreport}
OpenAI et~al. 2024.
\newblock \href {https://arxiv.org/abs/2303.08774} {Gpt-4 technical report}.
\newblock \emph{Preprint}, arXiv:2303.08774.

\bibitem[{Gu et~al.(2022)Gu, Zhang, Lee, Yoo, and Ha}]{gu-etal-2022-continuous}
Xiaodong Gu, Zhaowei Zhang, Sang-Woo Lee, Kang~Min Yoo, and Jung-Woo Ha. 2022.
\newblock \href {https://aclanthology.org/2022.coling-1.554/} {Continuous decomposition of granularity for neural paraphrase generation}.
\newblock In \emph{Proceedings of the 29th International Conference on Computational Linguistics}, pages 6369--6378, Gyeongju, Republic of Korea. International Committee on Computational Linguistics.

\bibitem[{Han et~al.(2024)Han, Zhang, Qi, Xu, Wang, Liu, Wang, Min, and Castelli}]{han2024ragqaarenaevaluatingdomain}
Rujun Han, Yuhao Zhang, Peng Qi, Yumo Xu, Jenyuan Wang, Lan Liu, William~Yang Wang, Bonan Min, and Vittorio Castelli. 2024.
\newblock \href {https://arxiv.org/abs/2407.13998} {Rag-qa arena: Evaluating domain robustness for long-form retrieval augmented question answering}.
\newblock \emph{Preprint}, arXiv:2407.13998.

\bibitem[{Ho et~al.(2020)Ho, Nguyen, Sugawara, and Aizawa}]{ho2020constructing}
Xanh Ho, Anh-Khoa~Duong Nguyen, Saku Sugawara, and Akiko Aizawa. 2020.
\newblock Constructing a multi-hop qa dataset for comprehensive evaluation of reasoning steps.
\newblock In \emph{Proceedings of the 28th International Conference on Computational Linguistics}, pages 6609--6625.

\bibitem[{Hsieh et~al.(2024)Hsieh, Chuang, Li, Wang, Le, Kumar, Glass, Ratner, Lee, Krishna, and Pfister}]{hsieh2024middlecalibratingpositionalattention}
Cheng-Yu Hsieh, Yung-Sung Chuang, Chun-Liang Li, Zifeng Wang, Long~T. Le, Abhishek Kumar, James Glass, Alexander Ratner, Chen-Yu Lee, Ranjay Krishna, and Tomas Pfister. 2024.
\newblock \href {https://arxiv.org/abs/2406.16008} {Found in the middle: Calibrating positional attention bias improves long context utilization}.
\newblock \emph{Preprint}, arXiv:2406.16008.

\bibitem[{Huang and Chang(2023)}]{huang2023reasoninglargelanguagemodels}
Jie Huang and Kevin Chen-Chuan Chang. 2023.
\newblock \href {https://arxiv.org/abs/2212.10403} {Towards reasoning in large language models: A survey}.
\newblock \emph{Preprint}, arXiv:2212.10403.

\bibitem[{Huang et~al.(2021)Huang, Cao, Parulian, Ji, and Wang}]{huang-etal-2021-efficient}
Luyang Huang, Shuyang Cao, Nikolaus Parulian, Heng Ji, and Lu~Wang. 2021.
\newblock \href {https://doi.org/10.18653/v1/2021.naacl-main.112} {Efficient attentions for long document summarization}.
\newblock In \emph{Proceedings of the 2021 Conference of the North American Chapter of the Association for Computational Linguistics: Human Language Technologies}, pages 1419--1436, Online. Association for Computational Linguistics.

\bibitem[{Jiang et~al.(2023)Jiang, Wu, Lin, Yang, and Qiu}]{jiang2023llmlinguacompressingpromptsaccelerated}
Huiqiang Jiang, Qianhui Wu, Chin-Yew Lin, Yuqing Yang, and Lili Qiu. 2023.
\newblock \href {https://arxiv.org/abs/2310.05736} {Llmlingua: Compressing prompts for accelerated inference of large language models}.
\newblock \emph{Preprint}, arXiv:2310.05736.

\bibitem[{Jiang et~al.(2024)Jiang, Wu, Luo, Li, Lin, Yang, and Qiu}]{jiang2024longllmlinguaacceleratingenhancingllms}
Huiqiang Jiang, Qianhui Wu, Xufang Luo, Dongsheng Li, Chin-Yew Lin, Yuqing Yang, and Lili Qiu. 2024.
\newblock \href {https://arxiv.org/abs/2310.06839} {Longllmlingua: Accelerating and enhancing llms in long context scenarios via prompt compression}.
\newblock \emph{Preprint}, arXiv:2310.06839.

\bibitem[{Jin et~al.(2025)Jin, Yu, Huang, Zeng, Wang, Hua, Zhao, Mei, Meng, Ding, Yang, Du, and Zhang}]{jin2025exploringconceptdepthlarge}
Mingyu Jin, Qinkai Yu, Jingyuan Huang, Qingcheng Zeng, Zhenting Wang, Wenyue Hua, Haiyan Zhao, Kai Mei, Yanda Meng, Kaize Ding, Fan Yang, Mengnan Du, and Yongfeng Zhang. 2025.
\newblock \href {https://arxiv.org/abs/2404.07066} {Exploring concept depth: How large language models acquire knowledge and concept at different layers?}
\newblock \emph{Preprint}, arXiv:2404.07066.

\bibitem[{Joshi et~al.(2017)Joshi, Choi, Weld, and Zettlemoyer}]{joshi2017triviaqa}
Mandar Joshi, Eunsol Choi, Daniel~S Weld, and Luke Zettlemoyer. 2017.
\newblock Triviaqa: A large scale distantly supervised challenge dataset for reading comprehension.
\newblock In \emph{Proceedings of the 55th Annual Meeting of the Association for Computational Linguistics (Volume 1: Long Papers)}, pages 1601--1611.

\bibitem[{Kuratov et~al.(2024)Kuratov, Bulatov, Anokhin, Rodkin, Sorokin, Sorokin, and Burtsev}]{kuratov2024babilongtestinglimitsllms}
Yuri Kuratov, Aydar Bulatov, Petr Anokhin, Ivan Rodkin, Dmitry Sorokin, Artyom Sorokin, and Mikhail Burtsev. 2024.
\newblock \href {https://arxiv.org/abs/2406.10149} {Babilong: Testing the limits of llms with long context reasoning-in-a-haystack}.
\newblock \emph{Preprint}, arXiv:2406.10149.

\bibitem[{Lewis et~al.(2021)Lewis, Perez, Piktus, Petroni, Karpukhin, Goyal, Küttler, Lewis, tau Yih, Rocktäschel, Riedel, and Kiela}]{lewis2021retrievalaugmentedgenerationknowledgeintensivenlp}
Patrick Lewis, Ethan Perez, Aleksandra Piktus, Fabio Petroni, Vladimir Karpukhin, Naman Goyal, Heinrich Küttler, Mike Lewis, Wen tau Yih, Tim Rocktäschel, Sebastian Riedel, and Douwe Kiela. 2021.
\newblock \href {https://arxiv.org/abs/2005.11401} {Retrieval-augmented generation for knowledge-intensive nlp tasks}.
\newblock \emph{Preprint}, arXiv:2005.11401.

\bibitem[{Li et~al.(2024)Li, Su, and Collier}]{li2024500xcompressorgeneralizedpromptcompression}
Zongqian Li, Yixuan Su, and Nigel Collier. 2024.
\newblock \href {https://arxiv.org/abs/2408.03094} {500xcompressor: Generalized prompt compression for large language models}.
\newblock \emph{Preprint}, arXiv:2408.03094.

\bibitem[{Liu et~al.(2024)Liu, Chen, Ji, Zhou, Chen, and Wang}]{liu2024rag}
Wanlong Liu, Junying Chen, Ke~Ji, Li~Zhou, Wenyu Chen, and Benyou Wang. 2024.
\newblock Rag-instruct: Boosting llms with diverse retrieval-augmented instructions.
\newblock \emph{arXiv preprint arXiv:2501.00353}.

\bibitem[{Mu et~al.(2024)Mu, Li, and Goodman}]{mu2024learningcompresspromptsgist}
Jesse Mu, Xiang~Lisa Li, and Noah Goodman. 2024.
\newblock \href {https://arxiv.org/abs/2304.08467} {Learning to compress prompts with gist tokens}.
\newblock \emph{Preprint}, arXiv:2304.08467.

\bibitem[{Pan et~al.(2024)Pan, Wu, Jiang, Xia, Luo, Zhang, Lin, Rühle, Yang, Lin, Zhao, Qiu, and Zhang}]{pan2024llmlingua2datadistillationefficient}
Zhuoshi Pan, Qianhui Wu, Huiqiang Jiang, Menglin Xia, Xufang Luo, Jue Zhang, Qingwei Lin, Victor Rühle, Yuqing Yang, Chin-Yew Lin, H.~Vicky Zhao, Lili Qiu, and Dongmei Zhang. 2024.
\newblock \href {https://arxiv.org/abs/2403.12968} {Llmlingua-2: Data distillation for efficient and faithful task-agnostic prompt compression}.
\newblock \emph{Preprint}, arXiv:2403.12968.

\bibitem[{Qiu et~al.(2025)Qiu, Shi, Zhao, Zhu, Zhang, and Feng}]{qiu2025latent}
Yilun Qiu, Tianhao Shi, Xiaoyan Zhao, Fengbin Zhu, Yang Zhang, and Fuli Feng. 2025.
\newblock Latent inter-user difference modeling for llm personalization.
\newblock \emph{arXiv preprint arXiv:2507.20849}.

\bibitem[{Que et~al.(2024)Que, Duan, He, Mou, Zhou, Liu, Rong, Wang, Yang, Zhang, Peng, Zhang, Zhang, and Chen}]{que2024hellobenchevaluatinglongtext}
Haoran Que, Feiyu Duan, Liqun He, Yutao Mou, Wangchunshu Zhou, Jiaheng Liu, Wenge Rong, Zekun~Moore Wang, Jian Yang, Ge~Zhang, Junran Peng, Zhaoxiang Zhang, Songyang Zhang, and Kai Chen. 2024.
\newblock \href {https://arxiv.org/abs/2409.16191} {Hellobench: Evaluating long text generation capabilities of large language models}.
\newblock \emph{Preprint}, arXiv:2409.16191.

\bibitem[{Rajpurkar et~al.(2016)Rajpurkar, Zhang, Lopyrev, and Liang}]{rajpurkar2016squad100000questionsmachine}
Pranav Rajpurkar, Jian Zhang, Konstantin Lopyrev, and Percy Liang. 2016.
\newblock \href {https://arxiv.org/abs/1606.05250} {Squad: 100,000+ questions for machine comprehension of text}.
\newblock \emph{Preprint}, arXiv:1606.05250.

\bibitem[{Shi et~al.(2023)Shi, Chen, Misra, Scales, Dohan, Chi, Schärli, and Zhou}]{shi2023largelanguagemodelseasily}
Freda Shi, Xinyun Chen, Kanishka Misra, Nathan Scales, David Dohan, Ed~Chi, Nathanael Schärli, and Denny Zhou. 2023.
\newblock \href {https://arxiv.org/abs/2302.00093} {Large language models can be easily distracted by irrelevant context}.
\newblock \emph{Preprint}, arXiv:2302.00093.

\bibitem[{Shi et~al.(2024)Shi, Zi, Shi, Zhang, Wu, and Xu}]{shi2024enhancingretrievalmanagingretrieval}
Yunxiao Shi, Xing Zi, Zijing Shi, Haimin Zhang, Qiang Wu, and Min Xu. 2024.
\newblock \href {https://arxiv.org/abs/2407.10670} {Enhancing retrieval and managing retrieval: A four-module synergy for improved quality and efficiency in rag systems}.
\newblock \emph{Preprint}, arXiv:2407.10670.

\bibitem[{Tarzanagh et~al.(2023)Tarzanagh, Li, Zhang, and Oymak}]{tarzanagh2023maxmargintokenselectionattention}
Davoud~Ataee Tarzanagh, Yingcong Li, Xuechen Zhang, and Samet Oymak. 2023.
\newblock \href {https://arxiv.org/abs/2306.13596} {Max-margin token selection in attention mechanism}.
\newblock \emph{Preprint}, arXiv:2306.13596.

\bibitem[{Vaswani et~al.(2023)Vaswani, Shazeer, Parmar, Uszkoreit, Jones, Gomez, Kaiser, and Polosukhin}]{vaswani2023attentionneed}
Ashish Vaswani, Noam Shazeer, Niki Parmar, Jakob Uszkoreit, Llion Jones, Aidan~N. Gomez, Lukasz Kaiser, and Illia Polosukhin. 2023.
\newblock \href {https://arxiv.org/abs/1706.03762} {Attention is all you need}.
\newblock \emph{Preprint}, arXiv:1706.03762.

\bibitem[{Verma(2024)}]{verma2024survey}
Sourav Verma. 2024.
\newblock Contextual compression in retrieval-augmented generation for large language models: A survey.
\newblock \emph{arXiv preprint arXiv:2409.13385}.

\bibitem[{Wang et~al.(2024)Wang, Chen, Cheng, Liao, Zhang, Wu, Yu, Xu, Zhang, Luo, Li, Yang, Huang, and Li}]{wang-etal-2024-leave}
Minzheng Wang, Longze Chen, Fu~Cheng, Shengyi Liao, Xinghua Zhang, Bingli Wu, Haiyang Yu, Nan Xu, Lei Zhang, Run Luo, Yunshui Li, Min Yang, Fei Huang, and Yongbin Li. 2024.
\newblock \href {https://doi.org/10.18653/v1/2024.emnlp-main.322} {Leave no document behind: Benchmarking long-context {LLM}s with extended multi-doc {QA}}.
\newblock In \emph{Proceedings of the 2024 Conference on Empirical Methods in Natural Language Processing}, pages 5627--5646, Miami, Florida, USA. Association for Computational Linguistics.

\bibitem[{Wang et~al.(2025{\natexlab{a}})Wang, Ding, Chen, Wu, Wang, Xie, and Zhao}]{wang2025vidorag}
Qiuchen Wang, Ruixue Ding, Zehui Chen, Weiqi Wu, Shihang Wang, Pengjun Xie, and Feng Zhao. 2025{\natexlab{a}}.
\newblock Vidorag: Visual document retrieval-augmented generation via dynamic iterative reasoning agents.
\newblock \emph{arXiv preprint arXiv:2502.18017}.

\bibitem[{Wang et~al.(2025{\natexlab{b}})Wang, Ding, Zeng, Chen, Chen, Wang, Xie, Huang, and Zhao}]{wang2025vrag}
Qiuchen Wang, Ruixue Ding, Yu~Zeng, Zehui Chen, Lin Chen, Shihang Wang, Pengjun Xie, Fei Huang, and Feng Zhao. 2025{\natexlab{b}}.
\newblock Vrag-rl: Empower vision-perception-based rag for visually rich information understanding via iterative reasoning with reinforcement learning.
\newblock \emph{arXiv preprint arXiv:2505.22019}.

\bibitem[{Wang et~al.(2023)Wang, Araki, Jiang, Parvez, and Neubig}]{wang2023learningfiltercontextretrievalaugmented}
Zhiruo Wang, Jun Araki, Zhengbao Jiang, Md~Rizwan Parvez, and Graham Neubig. 2023.
\newblock \href {https://arxiv.org/abs/2311.08377} {Learning to filter context for retrieval-augmented generation}.
\newblock \emph{Preprint}, arXiv:2311.08377.

\bibitem[{Xu et~al.(2023)Xu, Shi, and Choi}]{xu2023recompimprovingretrievalaugmentedlms}
Fangyuan Xu, Weijia Shi, and Eunsol Choi. 2023.
\newblock \href {https://arxiv.org/abs/2310.04408} {Recomp: Improving retrieval-augmented lms with compression and selective augmentation}.
\newblock \emph{Preprint}, arXiv:2310.04408.

\bibitem[{Yang et~al.(2025)Yang, Yang, Zhang, Hui, Zheng, Yu, Li, Liu, Huang, Wei, Lin, Yang, Tu, Zhang, Yang, Yang, Zhou, Lin, Dang, Lu, Bao, Yang, Yu, Li, Xue, Zhang, Zhu, Men, Lin, Li, Tang, Xia, Ren, Ren, Fan, Su, Zhang, Wan, Liu, Cui, Zhang, and Qiu}]{qwen2025qwen25technicalreport}
An~Yang, Baosong Yang, Beichen Zhang, Binyuan Hui, Bo~Zheng, Bowen Yu, Chengyuan Li, Dayiheng Liu, Fei Huang, Haoran Wei, Huan Lin, Jian Yang, Jianhong Tu, Jianwei Zhang, Jianxin Yang, Jiaxi Yang, Jingren Zhou, Junyang Lin, Kai Dang, Keming Lu, Keqin Bao, Kexin Yang, Le~Yu, Mei Li, Mingfeng Xue, Pei Zhang, Qin Zhu, Rui Men, Runji Lin, Tianhao Li, Tianyi Tang, Tingyu Xia, Xingzhang Ren, Xuancheng Ren, Yang Fan, Yang Su, Yichang Zhang, Yu~Wan, Yuqiong Liu, Zeyu Cui, Zhenru Zhang, and Zihan Qiu. 2025.
\newblock \href {https://arxiv.org/abs/2412.15115} {Qwen2.5 technical report}.
\newblock \emph{Preprint}, arXiv:2412.15115.

\bibitem[{Yang et~al.(2018)Yang, Qi, Zhang, Bengio, Cohen, Salakhutdinov, and Manning}]{yang2018hotpotqa}
Zhilin Yang, Peng Qi, Saizheng Zhang, Yoshua Bengio, William Cohen, Ruslan Salakhutdinov, and Christopher~D Manning. 2018.
\newblock Hotpotqa: A dataset for diverse, explainable multi-hop question answering.
\newblock In \emph{Proceedings of the 2018 Conference on Empirical Methods in Natural Language Processing}, pages 2369--2380.

\bibitem[{Zhu et~al.(2024)Zhu, Gu, Sikora, Ko, Liu, Lin, Shu, Luo, Meng, Liu, and Chen}]{zhu2024acceleratinginferenceretrievalaugmentedgeneration}
Yun Zhu, Jia-Chen Gu, Caitlin Sikora, Ho~Ko, Yinxiao Liu, Chu-Cheng Lin, Lei Shu, Liangchen Luo, Lei Meng, Bang Liu, and Jindong Chen. 2024.
\newblock \href {https://arxiv.org/abs/2405.16178} {Accelerating inference of retrieval-augmented generation via sparse context selection}.
\newblock \emph{Preprint}, arXiv:2405.16178.

\end{thebibliography}
\clearpage
\appendix
\section{Pseudocode}
The Pseudocode of \algname is provided in Algorithm~\ref{alg:attention_rag}. The algorithm takes a retrieved long context, query, and two language models as input. First, it generates an answer hint prefix based on the query to guide the attention mechanism. Then, it splits the long context into chunks of size m. For each chunk, it generates an anchor token using the compression model. If the anchor token is valid (not ``none''), it computes attention features using the anchor token and compresses the chunk accordingly. Finally, all compressed chunks are concatenated and used with the original query to generate the final answer.

\begin{algorithm}[ht]
\caption{Pseudocode of \algname.}
\label{alg:attention_rag}
\begin{algorithmic}[1]
\State \textbf{Input:} Retrieved long context \( C \), query \( q \), generation model \( L \), compression model \( L_C \)
\State \textbf{Output:} Generated sequence \( y \)
\State
\State \textbf{Generate Answer Hint Prefix}
\State Get answer hint prefix \( p \) through \( L \) with \( q \)
\State \textbf{Chunking}
\State Generate chunks \( c_1, \dots, c_n \) by partitioning \( C \) with chunk size \( m \), where \( n = \lceil |C| / m \rceil \)
\State Initialize empty variable \( C' \)
\State \textbf{Compressing with Attention}
\For{$j = 1$ to $n$}
    \State Generate the anchor token \( a_1 \) with \( L_C \), \( c_j \), \( q \), and \( p \)
    \If{$a_1$ is "none"}
        \State \textbf{continue}
    \Else
        \State Obtain Attention Features \( A_1 \) with \( a_1 \) and \( c_j \) \Comment{Eq. ~(\ref{eq:attention_feature})}
        \State Get compressed \( c_j' \) according to \( A_1 \) and \( c_j \) \Comment{Eq. ~(\ref{eq:compression_formula})}
        \State Append \( c_j' \) to \( C' \)
    \EndIf
\EndFor
\State \textbf{Generate} \( y \) from \( L \) with \( C' \) and \( q \)
\State \textbf{Return} Generated sequence \( y \)
\end{algorithmic}
\end{algorithm}

\section{Implementation Details}\label{sec:appendix}
\subsection{Prompt for generating answer prefix hint}\label{sec:app_hint}
We use the following prompt for generating answer prefix hint according to each query.
\begin{tcolorbox}[colback=gray!20,  colframe=gray!20, coltitle=black,  sharp corners, width=\linewidth, boxrule=0.5mm]
You are a formatting assistant. Given a question, your task is to generate a corresponding answering format. The format should maintain the same structure as the question but transform it into an incomplete answer template. If it is impossible to generate a format, return ``None''.
\\ \\
The format is like an complete answer, but truncated before the key word, and the key word is not included in the format.
\\ \\
For instance, if the question is ``Where is Daniel?'', the format should be ``Daniel is in the'', as the next word is the key word.
\\ \\
Note: For yes/no questions, such as ``Is Tom here?'', return ``None'' because these questions are typically answered with ``yes'' or ``no'' and do not have a natural continuation that leads to a single keyword.
\\ \\
Examples:

1. Question: Where is Daniel?

   Format: Daniel is in the
\vspace{8pt}\\
2. Question: What time is it?

   Format: It is
\vspace{8pt}\\
3. Question: Who is responsible for this?

   Format: The person responsible for this is
\vspace{8pt}\\
4. Question: Which film was released more recently, Dance With A Stranger or Miley Naa Miley Hum?

   Format: The film released more recently was
\vspace{8pt}\\
5. Question: Is Tom here?

   Format: None
\vspace{8pt}\\
In generation , you should only return the format, not any other text.

Now, here's a new question:

Question: {question}

Format:
\end{tcolorbox}
\subsection{Fixed Hint Prefix}\label{sec:app_fixed}
Our fixed hint prefix is: \textit{"Please output the most relevant keyword or phrase that is relevant to the answer of the question."}
\subsection{Prompt for generating anchor token}\label{subsec:attention_feature_prompt}
We use the following prompt to generate the anchor token for computing attention featrues.
\begin{tcolorbox}[colback=gray!20,  colframe=gray!20, coltitle=black,  sharp corners, width=\linewidth, boxrule=0.5mm]
You will be given a long context begin with 'Context:', a question begin with 'Question:', and a hint begin with 'Hint:'. Please answer the question.\\
Context: \{chunk\}\\
Hint: You should answer begin with \{prefix\_hint\}, if there is no useful information in the context for the question in the context and you really don't know the answer, just answer {prefix\_hint} none.\\
Question: \{question\}\\
Answer:\\
\{prefix\_hint\}
\end{tcolorbox}

\subsection{Hyperparameter settings}\label{subsec:hyperparameter_settings}

To reduce the randomness, we use greedy decoding in open-source LLMs generation. For the chunk size and $K$ in the attention-based compression process, we set them according to the context length in different benchmarks. In LongBench, where contexts are quite long, we use larger chunk size and $K$, in contrast, in BABILong, where we choose to experiment with mid-sized context, we use smaller chunk size and $K$. The detailed setting is shown in Table~\ref{tab:exp_hyperparameters}.
\begin{table}[h!]
\centering
\begin{tabular}{lcc}
\toprule
\textbf{Dataset} & \textbf{Chunk\_Size} & \textbf{$K$} \\
\midrule
HotpotQA      & 300 & 12 \\
2WikiMQA       & 300 & 15 \\
TriviaQA      & 150 & 8 \\
BABILong 1k   & 50  & 8  \\
BABILong 2k   & 100 & 10 \\
BABILong 4k   & 200 & 12 \\
\bottomrule
\end{tabular}
\caption{Hyperparameter settings of the experiment}
\label{tab:exp_hyperparameters}
\end{table}

\subsection{Attention Layer Choice}
In LLMs, attention layers capture different levels of information—shallow layers focus on syntax, while deeper ones encode semantics~\citep{benartzy2024attendfirstconsolidatelater,jin2025exploringconceptdepthlarge}. To fully exploit this hierarchy, we aggregate attention scores across all layers for compression, as detailed in~\secref{subsec:computing_attn}. This mitigates layer-specific bias and captures a broader information spectrum. We compare this approach with using shallow, middle, or deep layers alone on HotpotQA with Llama-3.1-8B-Instruct. 

As shown in Table~\ref{tab:layer_ablation}, full-layer aggregation yields superior performance, validating our strategy.

\begin{table}[t]
\centering
\setlength{\tabcolsep}{3.5pt}
\renewcommand{\arraystretch}{1}
\resizebox{0.8\linewidth}{!}{
\begin{tabular}{cccl}
\toprule
Layer Subsets & EM$\uparrow$ & LLM Judge $\uparrow$ &CR$\uparrow$ \\
\midrule
0 - 10  & 0.35 & 0.43 & \textbf{4.5x}\\
11 - 20& 0.38 & 0.50 & 3.6x\\
21 - 31 & 0.40 & 0.48 & 3.7x \\
\midrule
0 - 31 & \textbf{0.42} & \textbf{0.54} & 3.6x\\
\bottomrule
\end{tabular}
}
\caption{Performance on HotpotQA with different subset of layers}
\label{tab:layer_ablation}
\end{table}



\end{document}